\documentclass[lettersize,journal]{IEEEtran}
\usepackage{amsmath,amsfonts}
\usepackage{algorithmic}
\usepackage{algorithm}
\usepackage{array}
\usepackage{amssymb}
\usepackage{textcomp}
\usepackage{stfloats}
\usepackage{url}
\usepackage{bbm}
\usepackage{verbatim}
\usepackage[inline]{enumitem}
\usepackage{makecell}
\usepackage{booktabs}
\usepackage{siunitx}
\usepackage{multirow}
\usepackage{graphicx}
\usepackage{tabularray}
\usepackage{cite}
\usepackage[caption=false,font=footnotesize,labelfont=rm,textfont=rm,subrefformat=parens]{subfig}
\usepackage[colorlinks=true,linkcolor=red,citecolor=blue,urlcolor=blue,]{hyperref}
\renewcommand\ref[1]{\subref{#1}}
\begin{document}
\title{Dual-Stream Spectral Decoupling Distillation \\ for Remote Sensing Object Detection}

\author{Xiangyi Gao, Danpei Zhao*, \emph{Member, IEEE}, Bo Yuan, Wentao Li
	\thanks{Manuscript created Mar 20, 2025; revised July 22, 2025; accepted August 14, 2025.
		This work was supported by the National Natural Science Foundation of China under Grant 62271018 and in part by the Academic Excellence Foundation of BUAA for PhD Students. (Corresponding author: Danpei Zhao.)
		
		Xiangyi Gao, Danpei Zhao, Bo Yuan, and Wentao Li are with the Department of Aerospace Intelligent Science and Technology, School of Astronautics, Beihang University, Beijing 102206, China, and also with Key Laboratory of Spacecraft Design Optimization and Dynamic Simulation Technology, Ministry of Education (e-mail: gaoxiangyi23@buaa.edu.cn, zhaodanpei@buaa.edu.cn, yuanbobuaa@buaa.edu.cn, canoe@buaa.edu.cn).
		
	}}
\maketitle

\begin{abstract}
	Knowledge distillation is an effective and hardware-friendly method, which plays a key role in lightweighting remote sensing object detection.
	However, existing distillation methods often encounter the issue of mixed features in remote sensing images (RSIs), and neglect the discrepancies caused by subtle feature variations, leading to entangled knowledge confusion.
	To address these challenges, we propose an architecture-agnostic distillation method named Dual-Stream Spectral Decoupling Distillation (DS\textsuperscript{2}D\textsuperscript{2}) for universal remote sensing object detection tasks. Specifically, DS\textsuperscript{2}D\textsuperscript{2} integrates explicit and implicit distillation grounded in spectral decomposition. Firstly, the first-order wavelet transform is applied for spectral decomposition to preserve the critical spatial characteristics of RSIs. Leveraging this spatial preservation, a Density-Independent Scale Weight (DISW) is designed to address the challenges of dense and small object detection common in RSIs. Secondly, we show implicit knowledge hidden in subtle student-teacher feature discrepancies, which significantly influence predictions when activated by detection heads. This implicit knowledge is extracted via full-frequency and high-frequency amplifiers, which map feature differences to prediction deviations. Extensive experiments on DIOR and DOTA datasets validate the effectiveness of the proposed method. Specifically, on DIOR dataset, DS\textsuperscript{2}D\textsuperscript{2} achieves improvements of 4.2\% in AP\textsubscript{50} for RetinaNet and 3.8\% in AP\textsubscript{50} for Faster R-CNN, outperforming existing distillation approaches. The source code will be available at \url{https://github.com/PolarAid/DS2D2}.
\end{abstract}


\section{Introduction} 
\begin{figure}
		\centering
		\subfloat[Conventional Feature Distillation]
		{
			\includegraphics[width=0.48\textwidth]{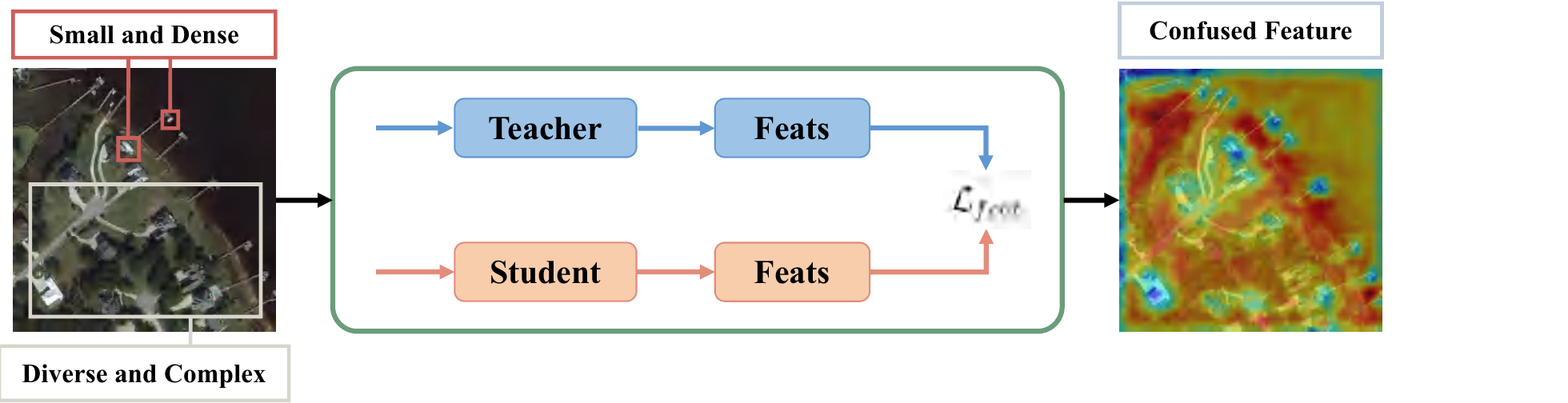}
			\label{Conventional}
		}
		
		\subfloat[Dual-Stream Spectral Decoupling Distillation (DS\textsuperscript{2}D\textsuperscript{2})]
		{
			\includegraphics[width=0.48\textwidth]{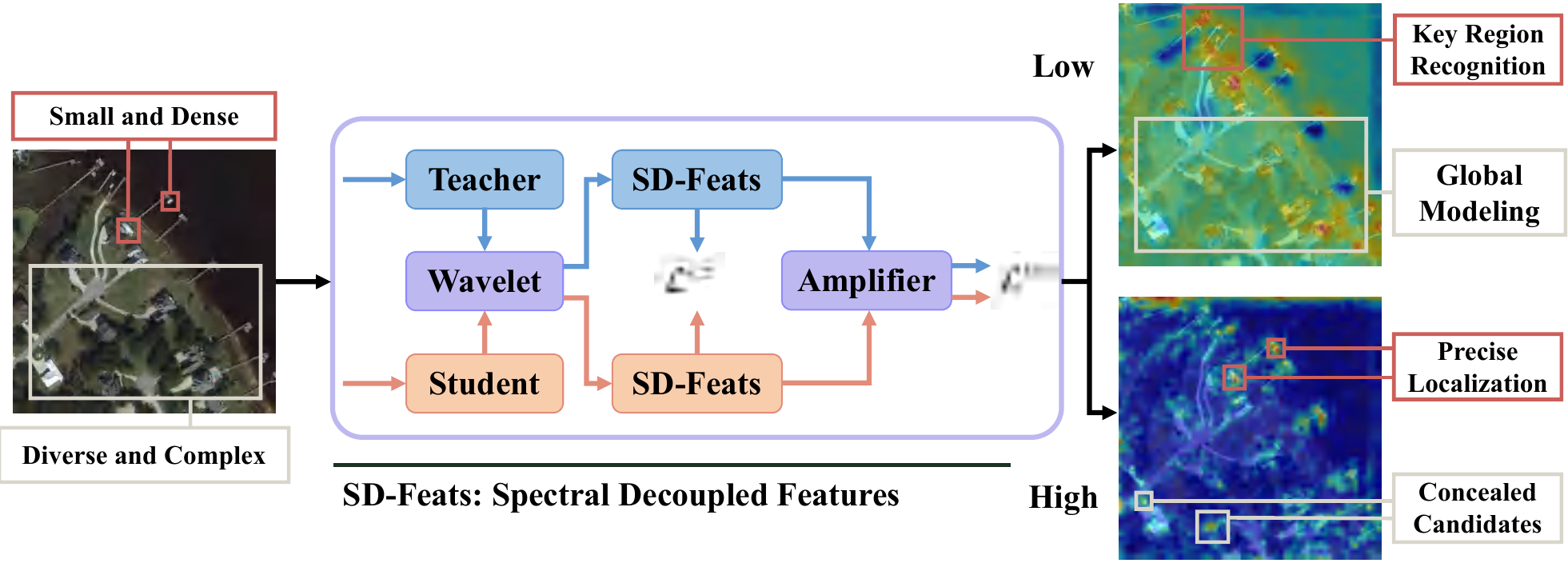}
			\label{Ours}
		}
		\caption{An overview of conventional feature distillation versus our DS\textsuperscript{2}D\textsuperscript{2}. Conventional methods struggle with semantic confusion and neglect implicit knowledge. We employ wavelet transforms for spectral decomposition. Besides, combining explicit and implicit distillation enables comprehensive learning.}
		\label{Intro}
	\end{figure}

	\IEEEPARstart{T}{he} rapid development of object detection algorithms has significantly enhanced information extraction capabilities in remote sensing images (RSIs). Existing advanced methods with complex structural designs \cite{ShipDetection}, \cite{InstanceSwitching-Based} suffer from slow inference speeds and face deployment challenges on hardware-constrained platforms. 
	As shown in Figure \ref{Intro}, RSIs typically cover diverse and complex scenes, where small objects are often obscured. This imposes additional challenges on detection methods, requiring more sophisticated discrimination and processing techniques to accurately extract critical features, thereby increasing the difficulty of lightweight optimization.
	To address the conflict between the massive streams of remote sensing data and the demand for rapid interpretation, numerous lightweight methods have been proposed.
	They primarily include efficient architecture design \cite{ANew}, \cite{CGC-NET}, \cite{LO-Det}, pruning \cite{Sparsity-AwareGlobal}, \cite{ALightweight}, quantization \cite{Data-Free}, \cite{GradQuant}, and knowledge distillation\cite{InsDist}, \cite{Two}. Among these, knowledge distillation has emerged as a dominant lightweight paradigm widely adopted in remote sensing tasks due to its deployment efficiency, robust performance, and hardware adaptability.

	Knowledge distillation was first proposed by Hinton in 2015 \cite{Dist}, and it has been extensively studied by researchers for remote sensing applications. 
	As a lightweight method, knowledge distillation does not reduce the computational cost of the model. It improves the model’s accuracy while keeping the computational cost unchanged. Therefore, the model’s computational efficiency is enhanced, achieving overall lightweighting.
	Related methods \cite{InsDist}, \cite{Two}, \cite{Adaptive}, \cite{Teaching} devise various feature-weighting and knowledge extraction mechanisms, yet overlook the inherent complexity and heterogeneity of remote sensing features, producing entangled knowledge representations.
	Other approaches \cite{Efficient}, \cite{TargetDetection}, \cite{Negative-Core}, \cite{LearningCross-Modality}, \cite{DPGD} enhanced feature transfer through multi-loss joint optimization, yet the loosely coupled relationships between losses often lead to conflicting gradients. 
	While methodological improvements to logits distillation continue to be explored \cite{Statistical}, the low-dimensional nature of logits creates an irreducible bottleneck for processing the immense information density inherent in RSIs.
	
	The intricate semantic complexity of RSIs poses significant challenges for object detection. Remote sensing feature maps exhibit exceptionally high information density, with individual pixels often encapsulating mixed features from multiple objects and complex backgrounds, resulting in erroneous feature combinations that exhibit high similarity in the embedding space. As shown in Fig. \ref{Intro}\subref{Conventional}, conventional methods directly utilize semantically confused feature maps for distillation, resulting in poor learning outcomes. Furthermore, while current approaches focus on explicit feature learning, unmodeled subtle distinctions frequently become performance bottlenecks.
	
	Based on the above analysis, we propose a Dual-Stream Spectral Decoupling Distillation framework. As shown in Fig. \ref{Intro}\subref{Ours}, the first-order wavelet transform is employed for spectral decomposition, offering three key advantages. 
	First, remote sensing mixed semantic information is separated into spectrally distinct components. Low-frequency components support key region recognition and global modeling, while high-frequency components are crucial for precise localization and the identification of concealed candidates. The wavelet transform also preserves directional texture details, effectively mitigating rotational viewpoint variations in remote sensing. 
	Second, spectral weights are adaptively redistributed and balanced. High-frequency information, often overwhelmed by low-frequency components, receives enhanced attention to prevent learning imbalance during model training. 
	Third, spatial information retention is achieved. Pure frequency-domain information cannot fully exploit the advantages of the space domain, such as detecting dense and small objects. To harmonize space-frequency domain strengths, the feature maps are decoupled into high/low-frequency components via the wavelet transform. Considering that multiple dense objects may aggregate at identical pixel positions in small-scale feature maps, the Density-Independent Scale Weight (DISW) is designed for optimization. DISW applies scale-specific weighting without being affected by object density. The essence of this approach remains direct feature map imitation between teacher and student models, hence termed explicit distillation.
	
	Furthermore, to address the failure of conventional distillation, we first analyze the causes, propose the concept of Knowledge Amplifier, and optimize it through spectral decomposition. It is revealed that subtle discrepancies in feature maps contain implicit knowledge, which is not adequately captured by conventional methods. This implicit knowledge significantly influences remote sensing detection outcomes when activated by detection heads. To resolve this issue, we investigate mechanisms for extracting implicit knowledge and formally define them as Knowledge Amplifiers. The teacher’s head constitutes an ideal Knowledge Amplifier, directly mapping feature map variations to prediction discrepancies and guiding model optimization. In explicit distillation, in which feature maps are decomposed into high-frequency and low-frequency components, corresponding Knowledge Amplifiers for both frequency components should be developed. Acquisition of these amplifiers involves retraining the teacher’s head, thereby incurring additional computational overhead. To minimize training costs, the existing teacher head is leveraged as a full-frequency amplifier to approximate the low-frequency counterpart, requiring only supplementary training of a high-frequency amplifier. 
	This distillation approach indirectly mimics features and can extract implicit knowledge overlooked by explicit feature distillation, hence being named implicit distillation. When the implicit distillation loss is optimized to zero, the teacher and student features become identical, although their final predictions may differ. Therefore, implicit distillation constitutes a form of feature distillation.
	
	To validate the superiority and robustness of DS\textsuperscript{2}D\textsuperscript{2}, extensive experiments are conducted on two large-scale remote sensing optical object detection datasets (DIOR\cite{DIOR} and DOTA\cite{DOTA}). The results demonstrate that DS\textsuperscript{2}D\textsuperscript{2} achieves superior performance compared to state-of-the-art methods. The key innovations and contributions of this work are summarized as follows:
	\begin{enumerate}
		\item We introduce spectral decomposition, utilizing the wavelet transform to address the information mixture in remote sensing feature maps. The decoupled features retain distinct spectral information while preserving critical spatial characteristics.
		
		\item To comprehensively learn decoupled features, the dual-stream distillation is designed for collaborative optimization. Explicit distillation directly mimics spatial-frequency domain knowledge, while implicit distillation employs knowledge amplifiers to excavate implicit patterns. The dual-stream distillation framework addresses the limitations of superficial and incomplete knowledge transfer.
		
		\item We propose a generic Dual-Stream Spectral Decoupling Distillation framework named DS\textsuperscript{2}D\textsuperscript{2} and validate its robust performance through extensive experiments on DIOR and DOTA datasets. Since DS\textsuperscript{2}D\textsuperscript{2} solely utilizes outputs from the neck and head, the algorithm demonstrates compatibility with diverse detection architectures. On DOTA dataset, DS\textsuperscript{2}D\textsuperscript{2} achieves a 4.3\% AP\textsubscript{50} improvement over the RetinaNet student.
	\end{enumerate}
\section{Related Work}
	\subsection{Object Detection in Aerial Images}
		Deep learning has emerged as the dominant paradigm in object detection. Typical CNN-based approaches are broadly categorized into single-stage and two-stage frameworks. Single-stage detectors\cite{SSD}, \cite{YOLO}, \cite{RetinaNet} directly predict bounding boxes and class probabilities in one pass, achieving efficient inference. In contrast, two-stage detectors\cite{FasterRCNN}, \cite{MaskRCNN} generate region proposals in the first stage and refine them for classification and localization in the second stage. Anchor-free methods \cite{FCOS}, \cite{RepPoints} eliminate predefined anchors by directly regressing object coordinates or key points, addressing the hyperparameter dependency of anchor-based approaches while introducing training challenges. Beyond CNN-based approaches, transformer-driven detectors\cite{DETR}, \cite{Swin} have gained traction for their end-to-end design and global context modeling capabilities. 
		However, these methods typically demand massive training datasets and struggle with small object detection, particularly in RSIs where small objects are prevalent.
		
		Most detection algorithms designed for natural images exhibit suboptimal performance on remote sensing data. This limitation primarily arises from complex semantics, dense object distributions, and extreme scale variations in such images. To overcome these challenges, a range of specialized approaches have been developed. RICNN\cite{RICNN} enhances rotation robustness by integrating rotation-invariant layers and feature consistency losses. Some methods\cite{Ship2018}, \cite{CSFF} introduce a cross-scale fusion framework to generate discriminative multi-level feature maps. Related approaches\cite{R2CNN}, \cite{ABNet} learn discriminative features via enhanced attention mechanisms, boosting object detection accuracy. Symmetric multimodal fusion is employed by SuperYOLO\cite{SuperYOLO} to extract complementary information, while a super-resolution branch is utilized to enhance high-resolution feature representations for small objects. 
		MaDiNet\cite{MaDiNet} employs a Gamma diffusion model to precisely estimate object scattering characteristics in Synthetic Aperture Radar (SAR) images, and incorporates an enhanced MambaSAR module to capture complex spatial dependencies of objects effectively.
		CFPT \cite{CFPT} designs cross-layer channelwise attention and cross-layer spatialwise attention to capture information. Additionally, CFPT incorporates global contextual information, which enhances the performance of small object detection.
		These methods explore diverse optimization strategies for remote sensing object detection, effectively addressing critical challenges and providing valuable insights for the design of distillation algorithms.
		
	\subsection{Knowledge Distillation for Remote Sensing Object Detection}
		Knowledge distillation has become a pivotal model compression technique for efficient deployment on resource-constrained devices, enabling lightweight models to achieve higher accuracy without compromising inference speed. Student models are trained by this approach to mimic ground-truth labels alongside the teacher model’s output distributions or intermediate features. Hinton et al.\cite{Dist} pioneered distillation by optimizing logits differences between teachers and students. FitNet\cite{FitNet} first introduced feature distillation via hidden-layer guidance. Established methodologies\cite{Mimicking}, \cite{Richness} enhance feature map distillation through region-aware weighting. Related methods \cite{FGD}, \cite{Global} incorporate attention mechanisms, prototypes, and relational features for collaborative distillation. Scholarly efforts \cite{Localization}, \cite{CrossKD} prioritize logits distillation, extracting essential knowledge via probability decoupling or hierarchical refinement. 
		
		Due to the significant discrepancy between remote sensing and natural images, suboptimal performance is often encountered when conventional distillation algorithms are directly applied. To overcome this limitation, several specialized distillation methods have been proposed. 
		ARSD\cite{Adaptive} and TWA\cite{Two} leverage multiscale knowledge in features, mitigating interference from object scale variations and background clutter in remote sensing object recognition. 
		InsDist\cite{InsDist} and AFD\cite{Efficient} investigate pixel-wise weight allocation in feature map distillation. Additionally, indirect optimization of feature imitation is achieved by distilling inter-feature relationships. 
		TDMD\cite{TargetDetection}, NSD\cite{Negative-Core}, and CMHRD\cite{LearningCross-Modality} design loss functions from multiple perspectives and levels for joint optimization. However, the weak correlations among these loss functions often lead to gradient conflicts.
		S\textsuperscript{3}MKM\cite{Statistical} takes into account the complexity of RSIs and designs a statistical sample selection module, which focuses on the learning of high-quality sample logits.
		DPGD\cite{DPGD} refines the prediction results and uses them to guide both relational distillation and logits distillation. However, the prediction results obtained under mixed features remain confusing, which diminishes their effectiveness in providing reliable guidance.
		Existing remote sensing distillation methods neglect the information mixture in feature maps, leading students to learn confused knowledge. Besides, the impact of subtle feature map discrepancies on prediction results is ignored by these methods, resulting in erroneous predictions even after feature map imitation.
	
	\subsection{Frequency Analysis in Deep Learning}
		Frequency-domain analysis, once prevalent in traditional methods, has seen limited integration with deep learning frameworks. Nevertheless, leveraging its inherent advantages, recent studies have successfully incorporated frequency-domain approaches into CNN architectures, achieving remarkable performance improvements. 
		ORSIm\cite{ORSIm} and SFS-Conv\cite{SFS} integrate spatial and frequency domains to optimize information-dense features. Some studies\cite{F3Net}, \cite{WaveletKD} uncover latent spatial information through spectral transforms and leverage their inherent properties to resolve challenges. FreeKD\cite{FreeKD} proposes a novel approach that generates a pixel-wise frequency mask by obtaining frequency prompts for weighted feature map distillation. However, overfitting may be induced by its overly complex processing mechanisms, resulting in compromised performance when applied to RSIs. The significant differences in spectral characteristics between remote sensing and natural images hinder the transferability of conventional methods to remote sensing scenarios. Furthermore, spatial features crucial for RSIs are often discarded by existing spectral analysis approaches, thereby leading to performance degradation.
	
	\begin{figure*}
		\centering
		\includegraphics[width=1.0\textwidth]{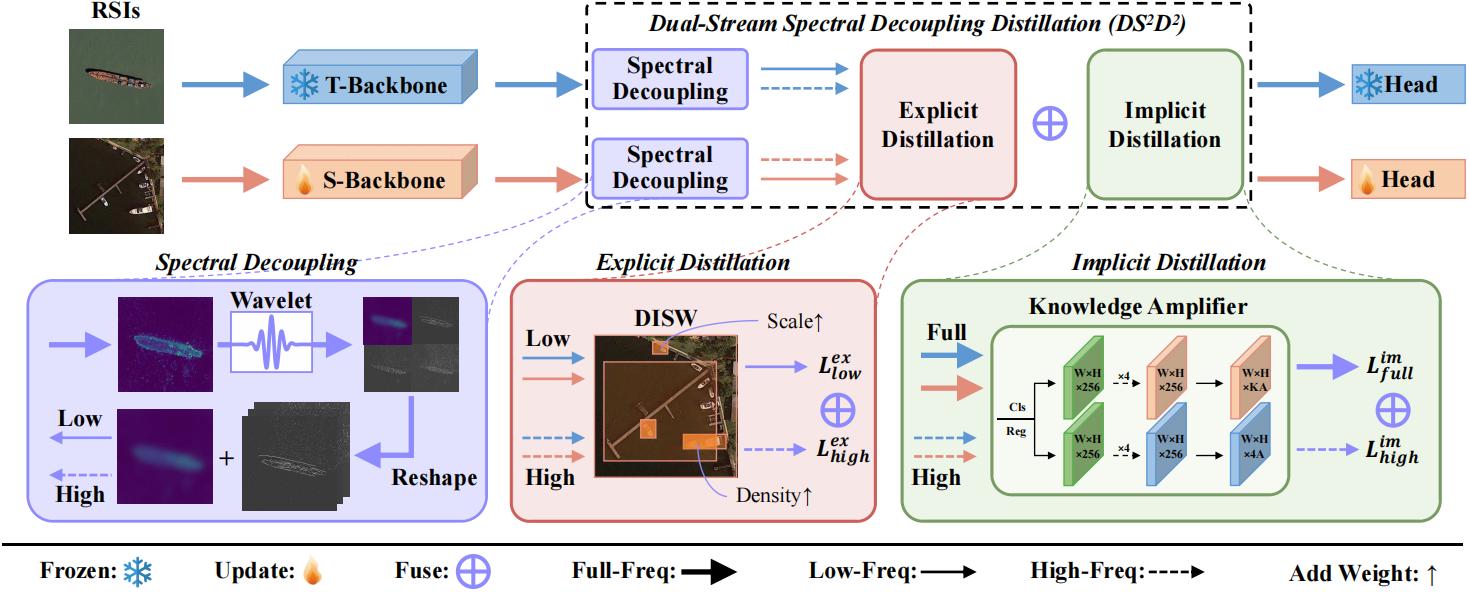}
		\caption{Overview of our DS\textsuperscript{2}D\textsuperscript{2}, with the knowledge amplifier structure exemplified by RetinaNet. Feature maps are generated from the input image through the networks. Feature maps are decoupled into high-frequency and low-frequency components via the wavelet transform to separate entangled semantic information in remote sensing. DISW applies weighting using spatial features preserved by the wavelet transform to compute the explicit distillation loss $L^{ex}$. To capture implicit knowledge, feature maps of different spectral components are fed into corresponding knowledge amplifiers to obtain predictions, which are then used to calculate the implicit distillation loss $L^{im}$.}
		\label{FlowChart}
	\end{figure*}

\section{Methodology}
	In this section, the principles and implementation details of the Dual-Stream Spectral Decoupling Distillation framework are elaborated, as illustrated in Fig. \ref{FlowChart}. We perform spectral decomposition of feature maps via the wavelet transform, separating valuable knowledge into spectral features. To enable thorough extraction and learning of the spectral knowledge, we develop a dedicated dual-stream distillation method.
	The decoupled feature maps are weighted by DISW to compute the explicit distillation loss. The original and high-frequency feature maps are subsequently fed into pretrained knowledge amplifiers, and the implicit distillation loss is calculated using the amplifiers’ output predictions.
	
	\subsection{Spectral Decomposition}
		Spectral decomposition is employed to address the issue of mixed information in remote sensing feature maps. To effectively handle multi-scale objects in RSIs, we utilize the multi-scale features from the Feature Pyramid Network (FPN) module for spectral decomposition and distillation. Since the batch size and multi-scale dimensions are independent of feature map computations, the feature maps can be simplified as $\mathcal{F}\in\mathbb{R}^{C\times H\times W}$. It has been demonstrated that the wavelet transform directly separates high-frequency and low-frequency features while effectively preserving spatial information. 
		Define the row-wise and column-wise low-pass filters as $h_0^{row}$ and $h_0^{col}$, and the row-wise and column-wise high-pass filters as $h_1^{row}$ and $h_1^{col}$. This paper employs the Haar filter, yielding
		\begin{equation}\label{eq0}
			\begin{aligned}
				&h_0^{col} = \frac{1}{\sqrt{2}}\left[\begin{matrix}
					1 \\
					1
				\end{matrix}\right]\quad
				&h_0^{row} = \frac{1}{\sqrt{2}}\left[\begin{matrix}
					1 & 1
				\end{matrix}\right]\quad\\
				&h_1^{col} = \frac{1}{\sqrt{2}}\left[\begin{matrix}
					1 \\
					-1
				\end{matrix}\right]
				&h_1^{row} = \frac{1}{\sqrt{2}}\left[\begin{matrix}
					1 & -1
				\end{matrix}\right]
			\end{aligned}.
		\end{equation}
		
		\begin{figure}
			\centering
			\includegraphics[width=0.45\textwidth]{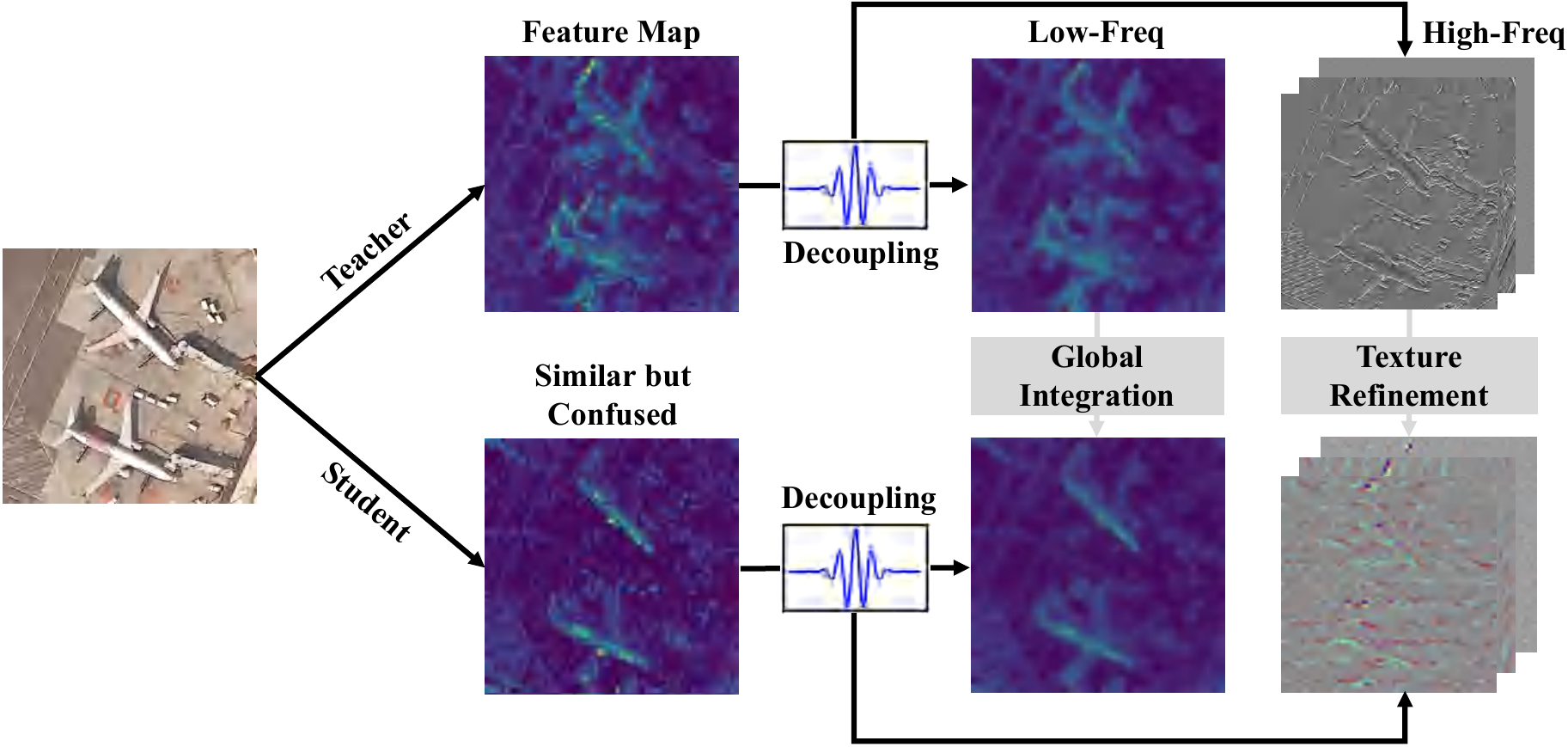}
			\caption{Schematic of spectral decomposition. Remote sensing feature maps exhibit vast and complex semantics, posing significant learning challenges. Spectral decomposition via the wavelet transform enables models to distinguish information effectively.}
			\label{Explicit}
		\end{figure}
		
		First, apply $h_0^{row}$ and $h_1^{row}$ to $\mathcal{F}$ for row filtering, obtaining $\mathcal{F}^r\in\mathbb{R}^{2C\times H\times\frac{W}{2}}$, where low-frequency and high-frequency components are concatenated along the channel dimension. Then, apply $h_0^{col}$ and $h_1^{col}$ to $\mathcal{F}^r$ for column filtering, yielding $\mathcal{F}^{rl}\in\mathbb{R}^{4C\times\frac{H}{2}\times\frac{W}{2}}$. Reshape $\mathcal{F}^{rl}$ to obtain $\mathcal{F}’\in\mathbb{R}^{C\times 4\times\frac{H}{2}\times\frac{W}{2}}$.
		
		Although wavelet transform inherently supports multi-scale feature extraction, the original feature maps already undergo multi-scale processing. Therefore, a first-order wavelet transform is adopted to maximize spatial information retention. Let the wavelet transform function be denoted as
		\begin{equation}\label{eq1}
			\mathcal{F}’ = \mathcal{W}(\mathcal{F})
		\end{equation}
		where $\mathcal{F}’$ can be decomposed into low-frequency components $\mathcal{L}$ and high-frequency components $\mathcal{H}$, i.e.,
		\begin{equation}\label{eq2}
			\mathcal{F}’ \rightarrow\left\{\begin{matrix}
				\mathcal{L} & \mathcal{L}\in\mathbb{R}^{C\times\frac{H}{2}\times\frac{W}{2}}\\
				\mathcal{H} & \mathcal{H}\in\mathbb{R}^{C\times3\times\frac{H}{2}\times\frac{W}{2}}\\
			\end{matrix}\right. .
		\end{equation}
		
	\subsection{Explicit Distillation}
		As shown in Fig. \ref{Explicit}, the feature maps of ambiguous semantics become more distinct after spectral decomposition, with more precise information and knowledge distribution across components to facilitate student learning. For instance, students acquire holistic learning of global information integration from low-frequency features while learning refined texture details from high-frequency features. The feature maps decoupled via the wavelet transform retain spatial characteristics, enabling direct utilization of these features for further processing. Small objects corresponding to fewer pixels are prone to be neglected during loss optimization. A classical approach involves designing the scale weighting to assign higher weights to small objects. Let the teacher feature map be denoted as $\mathcal{T}$, the student feature map as $\mathcal{S}$, the weight as $P$, and the loss as $L$. The basic formulation can be expressed as:
		\begin{equation}\label{eq3}
			L=\sum_{c=1}^C\sum_{h=1}^H\sum_{w=1}^WP_{c,h,w}*(\mathcal{T}_{c,h,w}-\mathcal{S}_{c,h,w})^2.
		\end{equation}
		
		In RSIs, dense objects are ubiquitous. This results in widespread object overlap within feature maps, where multiple small objects correspond to the same feature map pixel. We propose retaining contributions from all overlapping objects by summing their weights. For background pixels containing indispensable information, initial weights receive base allocations. Assuming there are $N$ objects in the image and the pixel set for each object is $D_i$, we define
		\begin{equation}\label{eq4}
			P_{c,h,w} = 1+\sum_{i=1}^N\frac{1}{|D_i|}* \mathbbm{1}_{\{(h,w)\in D_i\}}
		\end{equation}
		as the Density-Independent Scale Weight, where $\mathbbm{1}_{\{*\}}$ denotes the indicator function. This method addresses small-object challenges and prevents weight overlap issues caused by object aggregation.
		
		To verify the effectiveness of spectral decoupling, we used the identical form of loss function for features across different frequency bands. By performing explicit distillation on the disjoint spectral components separately, we obtain the explicit losses $L^{ex}$:
		\begin{equation}\label{eq5}
			\begin{aligned}
				L^{ex} &= \alpha L_{low}^{ex}+\beta L_{high}^{ex}\\
				L_{low}^{ex}&=\sum_{c=1}^C\sum_{h=1}^H\sum_{w=1}^WP_{c,h,w}\cdot(\mathcal{L}^\mathcal{T}_{c,h,w}-\mathcal{L}^\mathcal{S}_{c,h,w})^2\\
				L_{high}^{ex}&=\sum_{c=1}^C\sum_{h=1}^H\sum_{w=1}^WP_{c,h,w}\cdot(\mathcal{H}^\mathcal{T}_{c,h,w}-\mathcal{H}^\mathcal{S}_{c,h,w})^2
			\end{aligned}		
		\end{equation}
		where $L_{low}^{ex}$ and $L_{high}^{ex}$ denote the explicit distillation losses for low-frequency feature maps and high-frequency feature maps, respectively, and $\alpha$ and $\beta$ are the balancing hyperparameters.
		
	\subsection{Implicit Distillation}
		The information embedded in feature maps is rich and diverse, yet explicit distillation primarily extracts phenotypic knowledge. When the student’s feature maps closely resemble the teacher’s but yield inferior predictions, explicit distillation loses optimization direction. Therefore, we propose implicit distillation by designing knowledge amplifiers to guide student learning. Our knowledge amplifiers map subtle differences in feature maps to significant prediction discrepancies, akin to the detection head’s discriminative mechanism.
		
\begin{figure}
		 	\centering
		 	\includegraphics[width=0.48\textwidth]{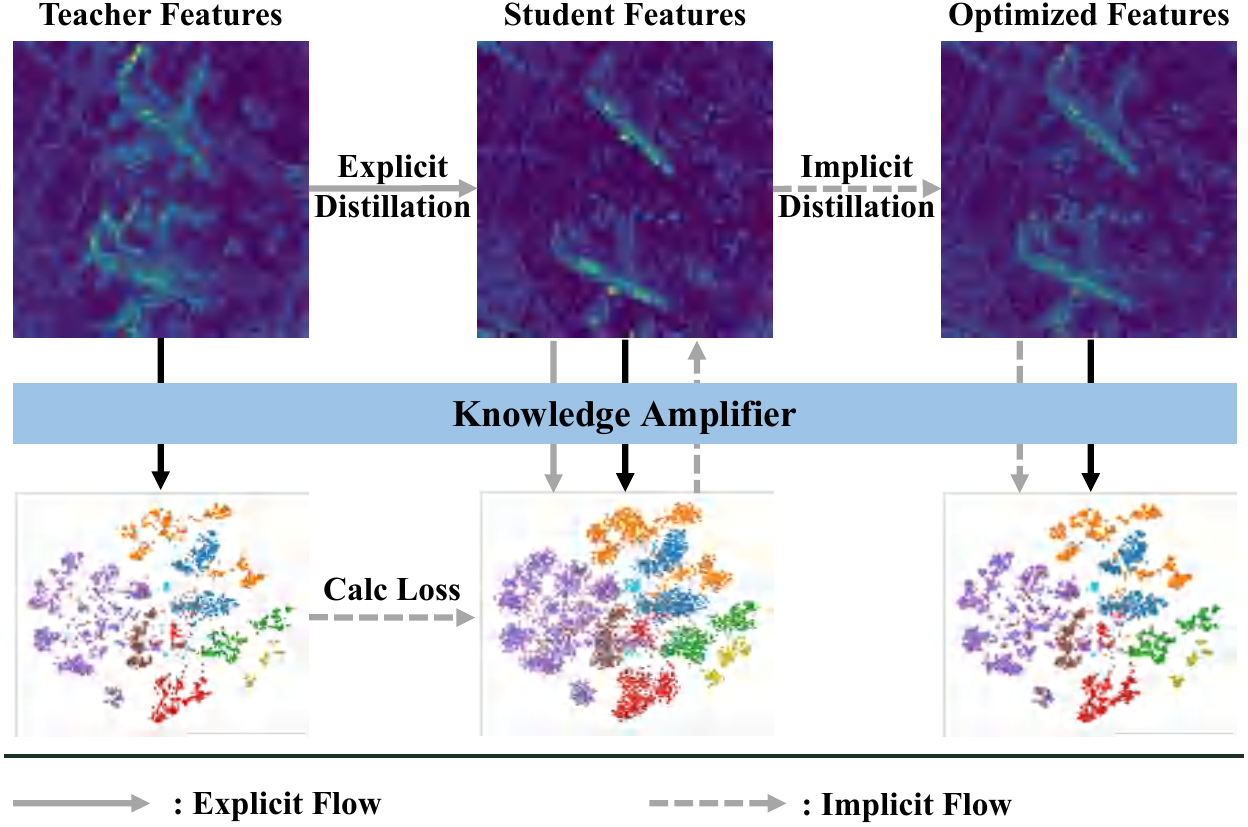}
		 	\caption{Schematic diagram of implicit distillation. Subtle discrepancies between teacher and student feature maps, processed by knowledge amplifiers, can significantly impact prediction results. Distilling these amplified prediction differences and leveraging backpropagation to optimize feature map learning enhances distillation performance.}
		 	\label{Implicit}
		\end{figure}

		Let the knowledge amplifier be denoted as $\mathcal{A}(\cdot)$, which shares the same architecture as the detection head. Detection heads typically consist of classification and regression branches. Thus, the feature map $\mathcal{F}$ processed by $\mathcal{A}(\cdot)$ yields $\mathcal{P}_{cls}$ and $\mathcal{P}_{reg}$. For teacher and student feature maps $\mathcal{T}$ and $\mathcal{S}$, their similarity may render explicit distillation ineffective for further improvement. However, through the mapping by $\mathcal{A}(\cdot)$, discrepancies can be amplified and leveraged to guide feature map learning via backpropagation. As illustrated in Fig. \ref{Implicit}, while the teacher-student model features demonstrate considerable similarity, notable discrepancies arise following the knowledge amplification process. Implicit distillation effectively mitigates these discrepancies.
		
		The feature maps are decoupled into high-frequency and low-frequency components in explicit distillation. Correspondingly, both components contain implicit knowledge. However, retraining knowledge amplifiers for low-frequency and high-frequency components requires additional computational resources. Thus, we directly replicate the teacher’s detection head weights as full-frequency knowledge amplifiers to approximate low-frequency knowledge amplifiers. The above process can be formulated as:  
		\begin{equation}\label{eq6}
			L_{full}^{im}=L_{reg}(\mathcal{P}_{reg}^{\mathcal{S}}, \mathcal{P}_{reg}^{\mathcal{T}})+L_{cls}(\mathcal{P}_{cls}^{\mathcal{S}}, \mathcal{P}_{cls}^{\mathcal{T}})
		\end{equation}
		where $L_{reg}$ and $L_{cls}$ are functions that compute prediction discrepancies. 
		 
		Additionally, we train high-frequency knowledge amplifiers using the teacher’s high-frequency feature maps $\mathcal{H}^\mathcal{T}$. 
		Given that object orientations vary widely within images, oriented high-frequency features at the image level cannot effectively guide learning. Therefore, high-frequency features from three orientations are directly fused.
		To align dimensions, we transform the high-frequency feature maps into ${\mathcal{H}^\mathcal{T}}’\in\mathbb{R}^{C\times H\times W}$:
		\begin{equation}\label{eq7}
			{\mathcal{H}^\mathcal{T}}’=\mathrm{Upsample}\left(\sum_{i=1}^{3}\mathcal{H}_{:, i, :, :}^{\mathcal{T}}\right).
		\end{equation}
		
		Let the high-frequency knowledge amplifier be denoted as $\mathcal{A}_{high}(\cdot)$. By feeding ${\mathcal{H}^\mathcal{T}}’$ and ${\mathcal{H}^\mathcal{S}}’$ into $\mathcal{A}_{high}(\cdot)$, we obtain $\mathcal{Q}_{cls}^{\mathcal{T}}$, $\mathcal{Q}_{reg}^{\mathcal{T}}$, $\mathcal{Q}_{cls}^{\mathcal{S}}$ and $\mathcal{Q}_{reg}^{\mathcal{S}}$. The high-frequency implicit distillation loss $L^{im}_{high}$ is then computed as:
		\begin{equation}\label{eq8}
			L^{im}_{high}=L_{reg}(\mathcal{Q}_{reg}^{\mathcal{S}}, \mathcal{Q}_{reg}^{\mathcal{T}})+L_{cls}(\mathcal{Q}_{cls}^{\mathcal{S}}, \mathcal{Q}_{cls}^{\mathcal{T}}).
		\end{equation}
		
		The implicit distillation loss is obtained by taking the weighted sum of the full-frequency implicit loss and the high-frequency implicit loss.
		\begin{equation}\label{eq9}
			L^{im} = \lambda L_{full}^{im} + \mu L_{high}^{im}
		\end{equation}
		where $\lambda$ and $\mu$ are the balancing hyperparameters.
		
	\subsection{Overall Objective}
		The total distillation loss is the sum of the explicit and implicit distillation losses, defined as $L_{dist}$, which is optimized jointly with the training loss:
		\begin{equation}\label{eq10}
			L_{dist} = L^{ex}+L^{im}.
		\end{equation}
		
\section{Experimental Results}
	In this section, we first introduce the datasets and evaluation metrics used in the experiments. Furthermore, the implementation details and experimental setup of the algorithm are elaborated. Subsequently, comparative experimental results are presented, and the strengths and weaknesses across methods are quantitatively analyzed. Finally, ablation studies are conducted to validate the effectiveness of the proposed algorithm.
	
	\subsection{Datasets}
		
		\begin{enumerate*}[font=\itshape, itemjoin=\\\hspace*{\parindent}]
			\item \textit{DIOR:} DIOR\cite{DIOR} is a large-scale optical remote sensing horizontal bounding box dataset released in 2020, containing 23,463 images with 800$\times$800 pixel dimensions. The dataset contains 20 object classes with 192,472 object instances. The official split divides the data into training, validation, and test sets. We combine the training and validation sets (11,725 images) for model training and report performance on the test set containing 11,738 images.
		\end{enumerate*}
		
		\begin{table}[tbp]
			\centering
			\caption{Comparative Experiments of Faster R-CNN on DIOR Dataset}\label{CMP-RCNN-DIOR}
			\begin{tblr}{
					hline{1,Z} = {1.5pt},
					colspec = {l|cccccc},
				}
				Method & mAP & AP$_{50}$ & AP$_{75}$ & AP$_S$ & AP$_M$ & AP$_L$\\
				\hline
				T: ResNet101 & 47.1 & 71.4 & 51.6 & 11.7 & 37.7 & 66.3\\
				S: ResNet18  & 41.9 & 67.1 & 45.0 & 10.1 & 32.4 & 60.0\\
				\hline
				FitNet\cite{FitNet} & 44.4 & 69.8 (2.7$\uparrow$) & 48.1 & 11.0 & 33.9 & 62.1\\
				FRS\cite{Richness} & 44.5 & 69.3 (2.2$\uparrow$) & 48.6 & 11.4 & 33.7 & 62.7\\
				FGD\cite{FGD} & 45.3 & 70.2 (3.1$\uparrow$) & 49.3 & 10.9 & 33.3 & \textbf{63.8}\\
				InsDist\cite{InsDist} & 44.1 & 69.8 (2.7$\uparrow$) & 47.8 & 11.6 & 34.0 & 61.5\\
				CrossKD\cite{CrossKD} & 44.5 & 69.1 (2.0$\uparrow$) & 48.4 & 11.3 & 33.7 & 63.1\\
				FreeKD\cite{FreeKD} & 44.8 & 69.9 (2.8$\uparrow$) & 48.6 & 11.5 & 33.0 & 62.6\\
				DPGD\cite{DPGD} & 44.5 & 69.7 (2.6$\uparrow$) & 48.4 & 11.2 & 34.0 & 62.2\\
				Ours & \textbf{45.6} & \textbf{70.9 (3.8$\uparrow$)} & \textbf{49.7} & \textbf{11.9} & \textbf{34.2} & 63.4\\
			\end{tblr}
		\end{table}
		
		\begin{table}[tbp]
			\centering
			\caption{Comparative Experiments of Faster R-CNN on DOTA Dataset}\label{CMP-RCNN-DOTA}
			\begin{tblr}{
					hline{1,Z} = {1.5pt},
					colspec = {l|cccccc},
				}
				Method & mAP & AP$_{50}$ & AP$_{75}$ & AP$_S$ & AP$_M$ & AP$_L$\\
				\hline
				T: ResNet101 & 40.9 & 63.8 & 44.8 & 17.6 & 42.0 & 54.6\\
				S: ResNet18  & 37.4 & 61.4 & 40.1 & 15.9 & 38.0 & 49.0\\
				\hline
				FitNet\cite{FitNet} & 38.1 & 62.1 (0.7$\uparrow$) & 41.3 & 16.5 & 38.5 & 50.1\\
				FRS\cite{Richness} & 39.0 & 62.5 (1.1$\uparrow$) & 42.2 & 16.9 & 39.5 & 50.3\\
				FGD\cite{FGD} & 38.7 & 62.0 (0.6$\uparrow$) & 42.1 & 17.1 & 39.3 & 51.2\\
				InsDist\cite{InsDist} & 38.1 & 62.0 (0.6$\uparrow$) & 41.3 & 17.1 & 39.1 & 49.8\\
				CrossKD\cite{CrossKD} & 39.0 & 62.4 (1.0$\uparrow$) & 42.4 & 16.5 & 39.5 & \textbf{51.9}\\
				FreeKD\cite{FreeKD} & 38.9 & 63.0 (1.6$\uparrow$) & \textbf{42.6} & \textbf{17.3} & 39.4 & 50.1\\
				DPGD\cite{DPGD} & 38.0 & 62.1 (0.7$\uparrow$) & 40.2 & 16.6 & 38.4 & 50.3\\
				Ours & \textbf{39.3} & \textbf{63.1(1.7$\uparrow$)} & \textbf{42.6} & 16.9 & \textbf{40.0} & 51.2\\
			\end{tblr}
		\end{table}
		
		\begin{enumerate*}[font=\itshape, itemjoin=\\\hspace*{\parindent}]
		\setcounter{enumi}{1}
			\item \textit{DOTA:} DOTA\cite{DOTA} is a large-scale optical remote sensing oriented bounding box dataset released in 2018. We convert the original oriented annotations into horizontal bounding boxes by extracting their minimum enclosing rectangles. The dataset contains 2806 images with approximate pixel dimensions of 4000$\times$4000 pixels. For computational efficiency, we split the images into $1024\times1024$ patches with 200-pixel overlaps. It contains 15 object categories with a total of 188,282 instances. The official partition provides training, validation, and test sets, but ground truth annotations are not publicly available for the test set. We train models on the training set and evaluate performance on the validation set.
		\end{enumerate*}
		
	\subsection{Implementation Details}
		All experiments are conducted on MMDetection\cite{MMDetection} v3.3.0 with PyTorch using four NVIDIA RTX 2080 Ti GPUs. Unless otherwise specified, ResNet101 and ResNet18 are used as the teacher and student backbones, respectively. We employ standard one-stage RetinaNet\cite{RetinaNet} and two-stage Faster R-CNN\cite{FasterRCNN} as baseline detectors, enabling systematic comparison of algorithmic performance across different detection paradigms. All models are optimized using SGD with a momentum of 0.9 for 24 epochs and a batch size of 8. Evaluation metrics include mean average precision (mAP) and its variants: AP\textsubscript{50}, AP\textsubscript{75}, AP\textsubscript{S}, AP\textsubscript{M}, and AP\textsubscript{L}.
		
		For RetinaNet, the learning rate is set to 0.005 with hyperparameters $\{\alpha=10^{-5}, \beta=10^{-5}, \lambda=1.0, \mu=1.0\}$ and $\{\alpha=7\times10^{-5}, \beta=5\times10^{-5}, \lambda=0.7, \mu=0.5\}$ for DIOR and DOTA datasets respectively. For Faster R-CNN, the learning rate is set to 0.02 with hyperparameters $\{\alpha=0.5, \beta=0.5, \lambda=10^{-5}, \mu=10^{-5}\}$ and $\{\alpha=0.12, \beta=1.2, \lambda=1.5\times10^{-3}, \mu=3\times10^{-3}\}$ applied to DIOR and DOTA datasets correspondingly.
		
		\begin{table}[tbp]
			\centering
			\caption{Comparative Experiments of RetinaNet on DIOR Dataset}\label{CMP-Reti-DIOR}
			\begin{tblr}{
					hline{1,Z} = {1.5pt},
					colspec = {l|cccccc},
				}
				Method & mAP & AP$_{50}$ & AP$_{75}$ & AP$_S$ & AP$_M$ & AP$_L$\\
				\hline
				T: ResNet101 & 45.0 & 68.4 & 48.3 & 8.6 & 35.1 & 64.4\\
				S: ResNet18  & 38.8 & 62.9 & 40.7 & 6.2 & 28.7 & 57.3\\
				\hline
				LogitsKD\cite{Dist} & 39.5 & 63.9 (1.0$\uparrow$)& 41.4 & 6.1 & 29.4 & 57.8\\
				FitNet\cite{FitNet} & 39.3 & 62.6 (0.3$\downarrow$)& 41.6 & 6.3 & 29.3 & 57.4\\
				FRS\cite{Richness} & 42.0 & 65.3 (2.4$\uparrow$)& 44.8 & 7.0 & 31.0 & 61.1\\
				FGD\cite{FGD} & 40.6 & 64.0 (1.1$\uparrow$)& 43.2 & 6.5 & 30.4 & 59.3\\
				InsDist\cite{InsDist} & 39.4 & 63.3 (0.4$\uparrow$) & 41.9 & 6.7 & 29.6 & 57.4\\
				CrossKD\cite{CrossKD} & 42.8 & 66.9 (4.0$\uparrow$)& 45.8 & 8.1 & 32.4 & 61.3\\
				FreeKD\cite{FreeKD} & 39.3 & 62.3 (0.6$\downarrow$)& 41.8 & 6.3 & 28.6 & 57.6\\
				DPGD\cite{DPGD} & 40.5 & 64.9 (2.0$\uparrow$)& 42.8 & 7.1 & 30.1 & 58.9\\
				Ours & \textbf{43.1} & \textbf{67.1 (4.2$\uparrow$)} & \textbf{45.9} & \textbf{8.6} & \textbf{33.3} & \textbf{61.5}\\
			\end{tblr}
		\end{table}
		
		\begin{table}[tbp]
			\centering
			\caption{Comparative Experiments of RetinaNet on DOTA Dataset}\label{CMP-Reti-DOTA}
			\begin{tblr}{
					hline{1,Z} = {1.5pt},
					colspec = {l|cccccc},
				}
				Method & mAP & AP$_{50}$ & AP$_{75}$ & AP$_S$ & AP$_M$ & AP$_L$\\
				\hline
				T: ResNet101 & 36.5 & 60.4 & 37.4 & 11.8 & 37.6 & 48.6\\
				S: ResNet18  & 31.2 & 55.2 & 30.8 & 10.0 & 32.4 & 41.0\\
				\hline
				LogitsKD\cite{Dist} & 31.9 & 55.7 (0.5$\uparrow$) & 31.8 & 9.6 & 33.4 & 41.8\\
				FitNet\cite{FitNet} & 30.7 & 53.7 (1.5$\downarrow$) & 30.7 & 9.4 & 32.3 & 40.1\\
				FRS\cite{Richness} & 33.4 & 57.1 (1.9$\uparrow$) & 33.8 & 10.9 & 34.4 & 45.5\\
				FGD\cite{FGD} & 32.4 & 55.4 (0.2$\uparrow$) & 32.8 & 11.2 & 33.5 & 44.0\\
				InsDist\cite{InsDist} & 31.0& 54.4 (0.8$\downarrow$) & 30.6 & 10.4 & 32.7 & 41.5\\
				CrossKD\cite{CrossKD} & 34.1 & 58.5 (3.3$\uparrow$) & 34.7 & \textbf{11.9} & 35.0 & 46.4\\
				FreeKD\cite{FreeKD} & 29.3 & 51.6 (3.6$\downarrow$) & 28.8 & 9.8 & 31.4 & 38.1\\
				DPGD\cite{DPGD} & 32.0 & 55.8 (0.6$\uparrow$) & 31.8 & 10.3 & 33.2 & 43.9\\
				Ours & \textbf{34.8} & \textbf{59.5 (4.3$\uparrow$)} & \textbf{35.5} & 11.8 & \textbf{36.0} & \textbf{47.3}\\
			\end{tblr}
		\end{table}
		
	\subsection{Comparison With State-of-the-Art Methods}
		To evaluate the effectiveness of the proposed method, DS\textsuperscript{2}D\textsuperscript{2} is compared with existing approaches including KD\cite{Dist}, FitNet\cite{FitNet}, FRS\cite{Richness}, FGD\cite{FGD}, InsDist\cite{InsDist}, CrossKD\cite{CrossKD}, FreeKD\cite{FreeKD}, and DPGD\cite{DPGD}. KD pioneers knowledge distillation by mimicking logits for image classification. FitNet introduces feature map distillation, achieving notable improvements in visual recognition tasks. FRS enhances object detection through pixel-level weighted optimization by computing feature richness. FGD employs attention-guided feature distillation and incorporates global feature relationships. InsDist extracts instance vectors and class prototypes from feature maps, and integrates feature distillation with relational distillation for comprehensive learning. CrossKD improves logits distillation via cross-head structures, while FreeKD employs frequency-domain masks to locate critical pixels for distillation. 
		DPGD leverages the models’ prediction results to optimize relational distillation and logits distillation.
		To ensure fair comparisons, all methods use the same pretrained teacher weights, student architectures, and training hyperparameters.
		
		\begin{figure*}
			\centering
			\includegraphics[width=0.95\textwidth]{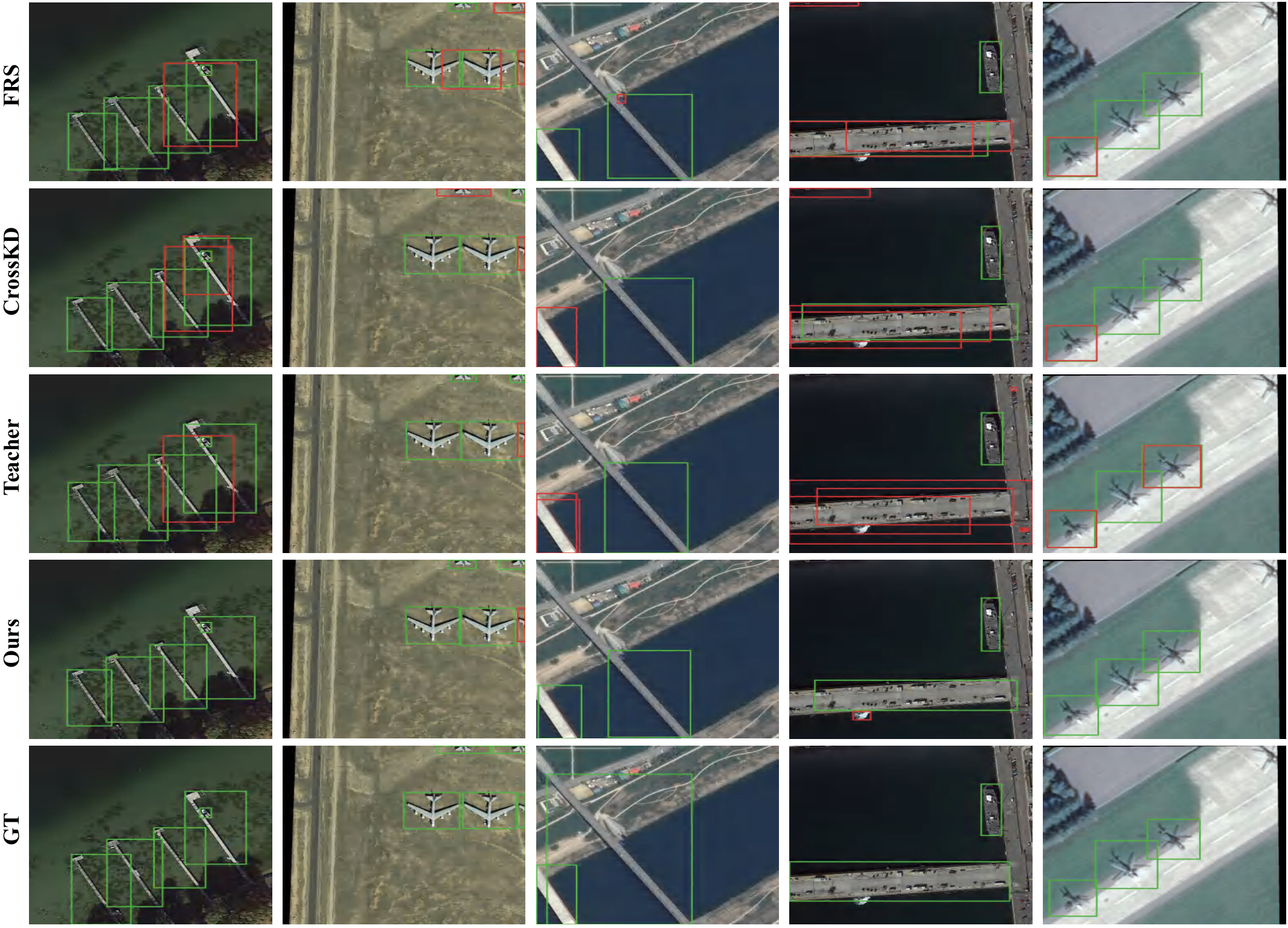}
			\caption{Visual detection results on DOTA dataset. Green and red boxes represent correct and incorrect predictions, respectively. Our DS\textsuperscript{2}D\textsuperscript{2} achieves higher detection accuracy, particularly for small and ambiguous objects, while surpassing the teacher model in specific scenarios.}
			\label{DOTA-Vis}
		\end{figure*}
	
		\begin{enumerate*}[font=\itshape, itemjoin=\\\hspace*{\parindent}]
			\item \textit{Comparative experiments on Faster R-CNN:} As shown in Table \ref{CMP-RCNN-DIOR} and Table \ref{CMP-RCNN-DOTA}, we deploy various distillation algorithms on Faster R-CNN and conduct experiments on DIOR and DOTA datasets. By leveraging spectral decomposition and knowledge amplifiers, DS\textsuperscript{2}D\textsuperscript{2} achieves the best performance. Specifically, DS\textsuperscript{2}D\textsuperscript{2} improves the student model by 3.8\% AP\textsubscript{50} on DIOR, outperforming the second-best method FGD by 0.7\% AP\textsubscript{50}. Notably, DS\textsuperscript{2}D\textsuperscript{2} significantly enhances small-object detection, with AP\textsubscript{S} surpassing that of the teacher model. The performance gaps among distillation methods are smaller on DOTA dataset due to the narrow performance gap between teacher and student models. Given that Faster R-CNN’s detection head primarily focuses on classification while remote sensing detection requires precise object localization and recognition, the cross-head module in CrossKD provides limited benefits.
		\end{enumerate*}
		
		\begin{table}[tbp]
			\centering
			\caption{Ablation Study on Explicit and Implicit Distillation}\label{Ablation-ExIm}
			\begin{tblr}{
					hline{1,Z} = {1.5pt},
					colspec = {cc|cccccc},
					colsep = 5pt,
				}
				Explicit & Implicit & mAP & AP$_{50}$ & AP$_{75}$ & AP$_S$ & AP$_M$ & AP$_L$\\
				\hline
				&  & 37.4 & 61.4 & 40.1 & 15.9 & 38.0 & 49.0 \\
				\checkmark &  & 38.4 & 62.7 & 41.4 & \textbf{17.2} & 39.0 & 50.0\\
				& \checkmark & 38.4 & 62.3 & 41.7 & 16.5 & 39.4 & 49.0 \\
				\checkmark & \checkmark & \textbf{39.3} & \textbf{63.1} & \textbf{42.6} & 16.9 & \textbf{40.0} & \textbf{51.2} \\ 
			\end{tblr}
		\end{table}
		
		\begin{enumerate*}[font=\itshape, itemjoin=\\\hspace*{\parindent}]
		\setcounter{enumi}{1}
			\item \textit{Comparative experiments on RetinaNet:} Comparative experimental results on RetinaNet are presented in Table \ref{CMP-Reti-DIOR} and Table \ref{CMP-Reti-DOTA}. The proposed DS\textsuperscript{2}D\textsuperscript{2} surpasses all competing methods, achieving 67.1\% AP\textsubscript{50} on the DIOR dataset, which represents a 4.2\% improvement over the student model. On the DOTA dataset, DS\textsuperscript{2}D\textsuperscript{2} outperforms the second-best method CrossKD by 1.0\% AP\textsubscript{50}. InsDist, designed for RSIs, fails to account for the information mixture in feature maps, resulting in confused instance vectors and class prototypes that degrade distillation performance. DS\textsuperscript{2}D\textsuperscript{2} effectively transfers small-object knowledge, enabling the student model to match the teacher’s AP\textsubscript{S} on both datasets. 
			FreeKD delivers suboptimal performance on RSIs, as it essentially remains an explicit spectral feature weighting method for distillation. This leads to inadequate learning of spectral knowledge. Moreover, FreeKD fails to leverage the spatial advantages of wavelet transforms, discarding substantial spatial domain knowledge.
		\end{enumerate*}
	
		\begin{figure}
			\centering
			\subfloat[InsDist]
			{
				\includegraphics[width=0.23\textwidth]{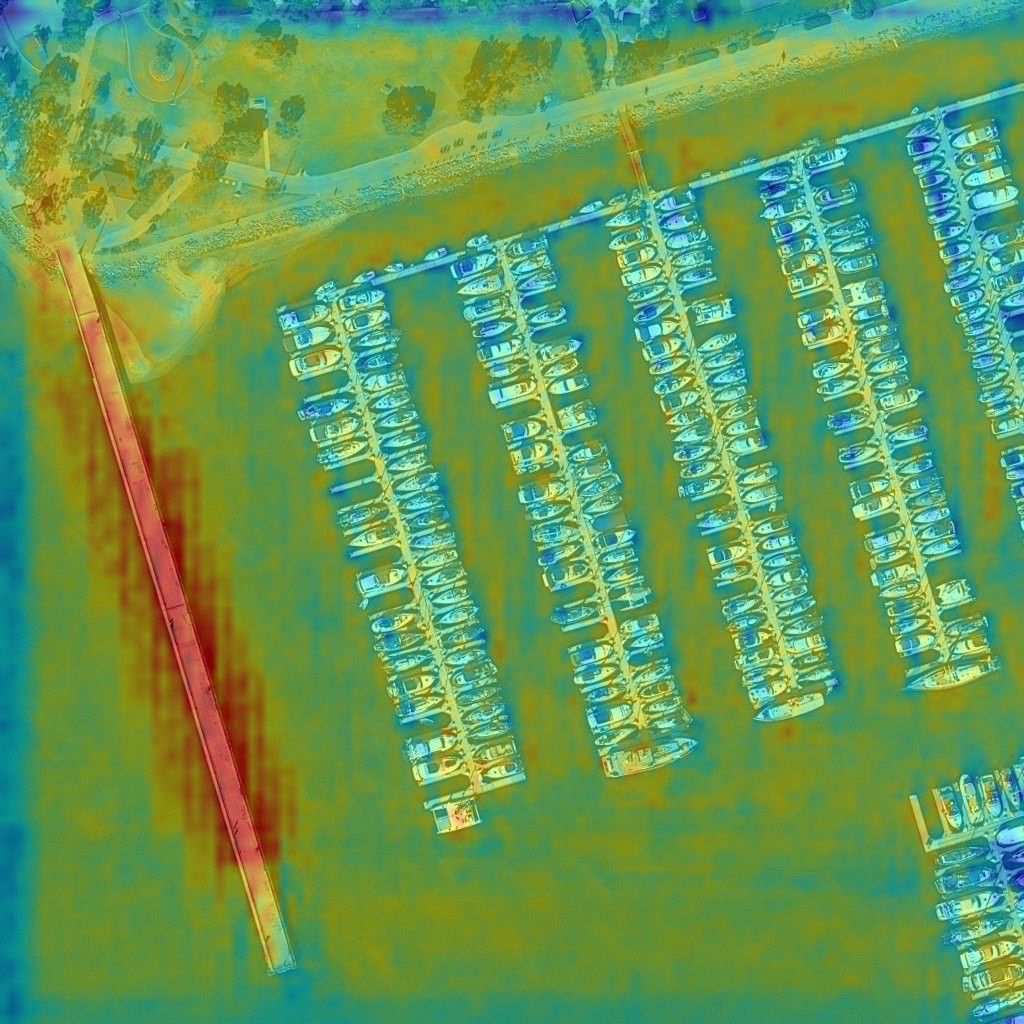}
				\label{HeatMap-InsDist}
			}
			\subfloat[DS\textsuperscript{2}D\textsuperscript{2}]
			{
				\includegraphics[width=0.23\textwidth]{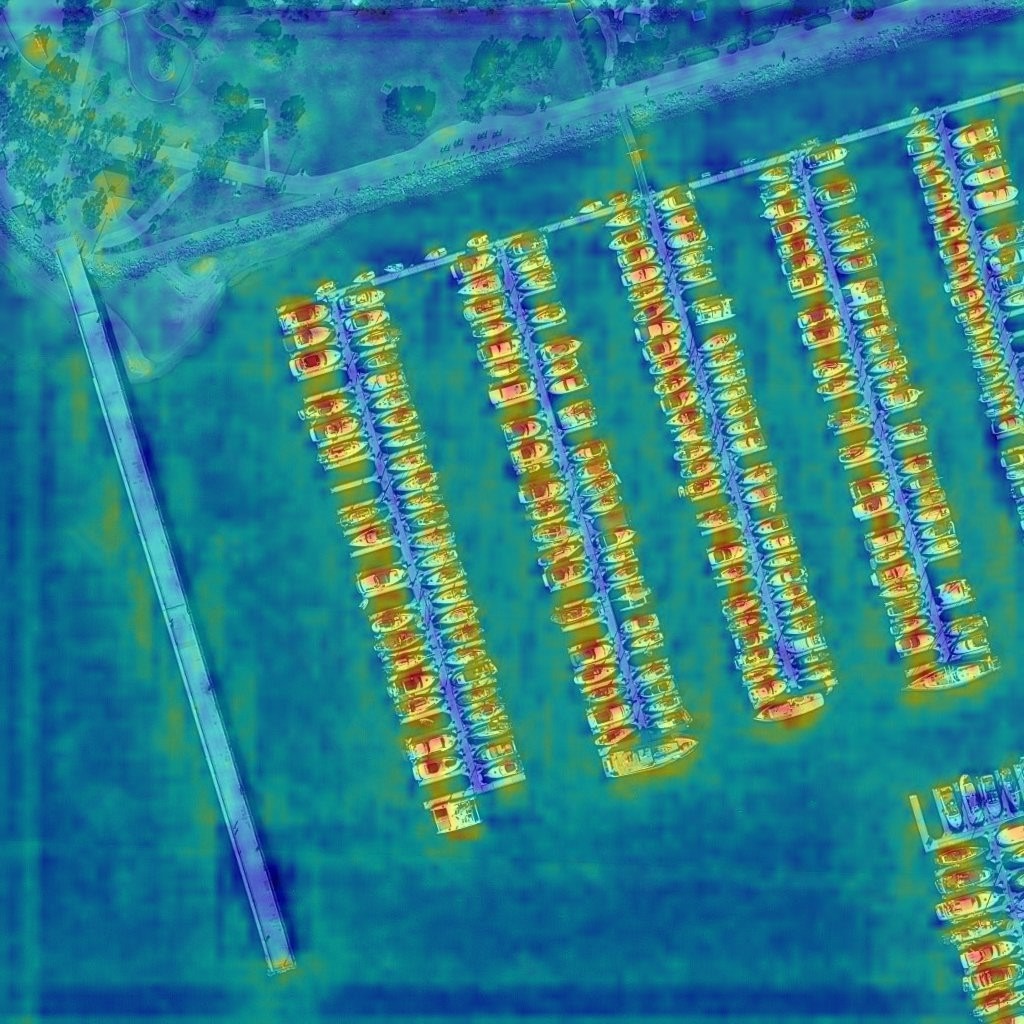}
				\label{HeatMap-D3}
			}
			\caption{Heatmap visualization on DOTA dataset. (a) corresponds to InsDist, and (b) corresponds to DS\textsuperscript{2}D\textsuperscript{2}. Experimental results demonstrate that DS\textsuperscript{2}D\textsuperscript{2} effectively detects dense and small objects while exhibiting enhanced false-positive suppression capability.}
			\label{HeatMap}
		\end{figure}
		
		\begin{table}[tbp]
			\centering
			\caption{Ablation Study on Spectral Decomposition}\label{Ablation-LowHigh}
			\begin{tblr}{
					hline{1,Z} = {1.5pt},
					colspec = {c|c|cc|ccc},
					colsep = 5pt,
				}
				Network & Dataset & Low & High & mAP & AP$_{50}$ & AP$_{75}$\\
				\hline
				\SetCell[r=8]{m} Faster R-CNN & \SetCell[r=4]{m} DIOR &  &  & 41.9 & 67.1 & 45.0 \\
				&  & \checkmark &  & 45.3 & 70.2 & \textbf{49.7} \\
				&  &  & \checkmark & 43.5 & 68.7 & 47.4 \\
				&  & \checkmark & \checkmark & \textbf{45.6} & \textbf{70.9} & \textbf{49.7} \\
				\cline{2-7}
				& \SetCell[r=8]{m} DOTA &  &  & 37.4 & 61.4 & 40.1 \\
				&  & \checkmark &  & 37.8 & 61.9 & 40.2 \\
				&  &  & \checkmark & 38.6 & 62.3 & 41.8 \\
				&  & \checkmark & \checkmark & \textbf{39.3} & \textbf{63.1} & \textbf{42.6} \\
				\cline{1-1} \cline{3-7}
				\SetCell[r=4]{m} RetinaNet &  &  &  & 31.2 & 55.2 & 30.8 \\
				&  & \checkmark &  & 33.9 & 58.0 & 34.0 \\
				&  &  & \checkmark & 33.9 & 57.4 & 34.7 \\
				&  & \checkmark & \checkmark & \textbf{34.8} & \textbf{59.5} & \textbf{35.5} \\
			\end{tblr}
		\end{table}
		
		\begin{enumerate*}[font=\itshape, itemjoin=\\\hspace*{\parindent}]
		\setcounter{enumi}{2}
			\item \textit{Visual Comparison of Algorithm Performance:} For qualitative algorithm comparison, the detection results of multiple methods are visualized. Fig. \ref{DOTA-Vis} shows detection results from RetinaNet with the DOTA dataset, including DS\textsuperscript{2}D\textsuperscript{2}, CrossKD, FRS, teacher predictions, and the ground truth annotations. The results demonstrate DS\textsuperscript{2}D\textsuperscript{2}’s superior capability in managing overlapping bounding boxes (harbors in Column 1) and partial object detection (planes in Column 2). By leveraging frequency-domain information extraction, DS\textsuperscript{2}D\textsuperscript{2} effectively distinguishes semantically ambiguous objects (Columns 4-5). The distilled model achieves performance comparable to the teacher model, even demonstrating occasional performance advantages in specific scenarios.\\
			
			\hspace*{\parindent}We extract features distilled by InsDist and DS\textsuperscript{2}D\textsuperscript{2} respectively, generating heatmaps for visual analysis. Experimental results are shown in Fig. \ref{HeatMap}. Compared with InsDist, DS\textsuperscript{2}D\textsuperscript{2} demonstrates stronger background feature suppression capability while achieving more precise detection of dense and small objects. This validates that spectral decomposition combined with DISW learning enhances DS\textsuperscript{2}D\textsuperscript{2}’s semantic discriminative power in remote sensing scenarios.
			
		\end{enumerate*}
		
	\subsection{Ablation Studies}
		In this section, we conduct ablation studies to validate the effectiveness of each component in the proposed algorithm.
		
		\begin{enumerate*}[font=\itshape, itemjoin=\\\hspace*{\parindent}]
			\item \textit{Dual-Stream Distillation Impact Analysis:} Experiments demonstrate that significant performance gaps persist despite close alignment via direct imitation between student and teacher feature maps. This is attributed to subtle discrepancies in student feature maps containing implicit knowledge. Such knowledge is activated through detection heads, substantially impacting predictions. To address this, we propose knowledge amplifiers and implicit distillation, and integrate them with explicit distillation to form dual-stream distillation. The performance is detailed in Table \ref{Ablation-ExIm}. Using Faster R-CNN with the DOTA dataset, standalone explicit and implicit distillation achieve 1.3\% and 0.9\% AP$_{50}$ improvements, respectively. The combined implementation achieves a 1.7\% AP$_{50}$ improvement.
		\end{enumerate*}
		
		\begin{table}[tbp]
			\centering
			\caption{Ablation Study of DISW on DIOR Dataset}\label{Ablation-DISW}
			\begin{tblr}{
					hline{1,Z} = {1.5pt},
					colspec = {c|cccccc},
				}
				DISW & mAP & AP$_{50}$ & AP$_{75}$ & AP$_S$ & AP$_M$ & AP$_L$\\
				\hline
				& 45.3 & 70.4 & 49.5 & 11.5 & 33.9 & \textbf{63.4}\\
				\checkmark & \textbf{45.6} & \textbf{70.9} & \textbf{49.7} & \textbf{11.9} & \textbf{34.2} & \textbf{63.4}\\
			\end{tblr}
		\end{table}
		
		\begin{table}[tbp]
			\centering
			\caption{Ablation Study on Frequency-Domain Processing Methods}\label{Ablation-Wavelet}
			\begin{tblr}{
					hline{1,Z} = {1.5pt},
					colspec = {c|ccccccc},
					colsep = 5pt,
				}
				Dataset & Method & mAP & AP$_{50}$ & AP$_{75}$ & AP$_S$ & AP$_M$ & AP$_L$\\
				\hline
				\SetCell[r=4]{m} DIOR & Teacher & 45.0 & 68.4 & 48.3 & 8.6 & 35.1 & 64.4\\
				& Student & 38.8 & 62.9 & 40.7 & 6.2 & 28.7 & 57.3\\
				\cline{2-8}
				& FFT & 42.8 & 66.4 & 45.7 & 8.1 & 32.3 & 61.4\\
				& Wavelet & \textbf{43.1} & \textbf{67.1} & \textbf{45.9} & \textbf{8.6} & \textbf{33.3} & \textbf{61.5}\\
				\hline
				\SetCell[r=4]{m} DOTA & Teacher & 36.5 & 60.4 & 37.4 & 11.8 & 37.6 & 48.6\\
				& Student & 31.2 & 55.2 & 30.8 & 10.0 & 32.4 & 41.0\\
				\cline{2-8}
				& FFT & 34.4 & 58.6 & 35.1 & \textbf{12.2} & \textbf{36.0} & 46.5\\
				& Wavelet & \textbf{34.8} & \textbf{59.5} & \textbf{35.5} & 11.8 & \textbf{36.0} & \textbf{47.3}\\
			\end{tblr}
		\end{table}
		
		\begin{enumerate*}[font=\itshape, itemjoin=\\\hspace*{\parindent}]
			\setcounter{enumi}{1}
			\item \textit{Analysis of Spectral Impact:} We propose to decouple feature maps into low-frequency and high-frequency components to address semantic aliasing in remote sensing feature maps. To validate this approach, ablation studies are conducted (Table \ref{Ablation-LowHigh}). Results show noticeable performance improvements when distilling with low-frequency or high-frequency feature maps alone. Combining both components during distillation further enhances the performance to optimal levels. For example, when low-frequency and high-frequency distillation are separately applied to RetinaNet, the AP$_{50}$ on DOTA improves by 2.8\% and 2.2\%, respectively. Combined implementation demonstrates a 4.3\% AP\textsubscript{50} improvement.
		
		\end{enumerate*}
		
		\begin{enumerate*}[font=\itshape, itemjoin=\\\hspace*{\parindent}]
			\setcounter{enumi}{2}		
			\item \textit{Performance Impact of DISW:} Existing research extensively explores spatial-domain processing of feature maps due to the distinctive spatial characteristics of remote sensing objects. However, frequency-domain processing struggles to utilize these characteristics fully. Spectral decomposition is employed to preserve spatial properties during frequency-domain operations. To tackle dense and small object challenges in RSIs, we design the DISW module for optimization. Experimental results in Table \ref{Ablation-DISW} demonstrate that a 0.5\% AP$_{50}$ improvement is achieved with DISW, thereby validating its effectiveness. DISW may produce larger weights in the dense small object regions of small-sized feature maps, but the total loss is not excessively large due to the fewer pixels. When optimizing feature maps of different sizes simultaneously, the gradients of larger feature maps are smoother and dominate. Additionally, the sparse data within the same batch will dilute the gradients generated by the dense objects.
		\end{enumerate*}
		
		\begin{figure}
			\centering
			\includegraphics[width=0.45\textwidth]{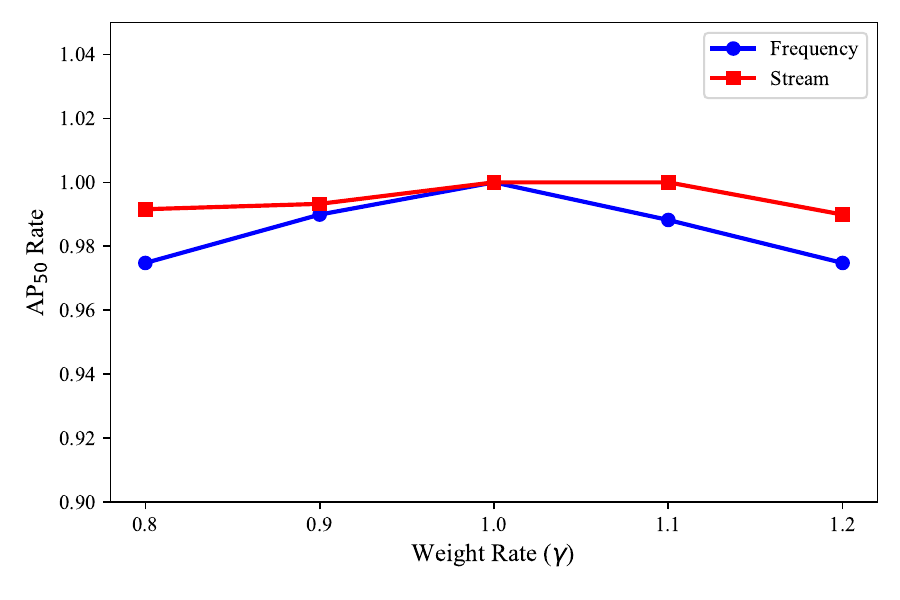}
			\caption{Hyperparameter tuning experimental results. The impact of both spectral balance and stream balance on performance is examined. The x-axis represents balancing factor values, and the y-axis indicates the current-to-initial AP$_{50}$ ratio. Blue curves represent spectral balance effects, while red curves indicate stream balance influences.}
			\label{LinePlot}
		\end{figure}
		
		\begin{enumerate*}[font=\itshape, itemjoin=\\\hspace*{\parindent}]
		\setcounter{enumi}{3}		
			\item \textit{Impact of Frequency Analysis Methods:} Selection criteria must be tailored to the characteristics of RSIs. Classical FFT provides only global frequency information, failing to capture object edges and textures, whereas the wavelet transform achieves this. Additionally, the wavelet transform extracts directional features through basis function selection, which is highly advantageous for remote sensing viewing angles. Given the inherent multi-scale information in feature maps, the first-order wavelet transform is employed to preserve spatial detail. Experimental results (Table \ref{Ablation-Wavelet}) demonstrate the superiority of the wavelet transform over FFT, with AP$_{50}$ improving by 0.7\% on DIOR and 0.9\% on DOTA. This validates the dual capacity of the wavelet transform to capture both frequency and spatial features, thereby demonstrating its effectiveness in complex scenarios.
			
		\end{enumerate*}
		
		\begin{figure}
			\centering
			\includegraphics[width=0.48\textwidth]{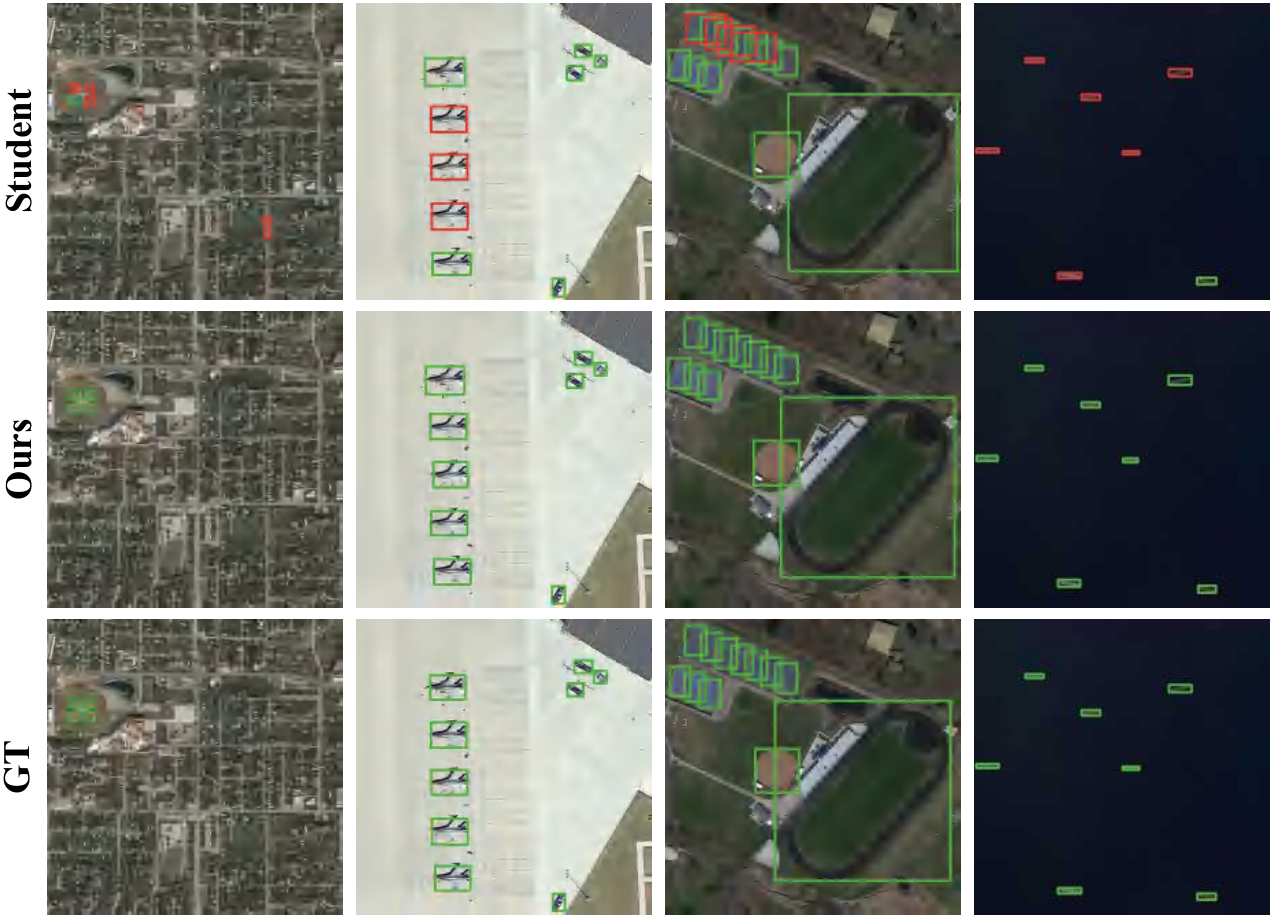}
			\caption{Visual detection results before and after distillation on DIOR dataset. Green boxes denote correct predictions. Red boxes denote incorrect predictions. The model exhibits significant performance improvement after distillation, achieving detection accuracy on par with ground truths across diverse scenarios.}
			\label{DIOR-Vis}
		\end{figure}
		
		\begin{enumerate*}[font=\itshape, itemjoin=\\\hspace*{\parindent}]
		\setcounter{enumi}{4}		
			\item \textit{Performance Analysis under Stronger Teacher/Student:} We investigate the impact of stronger teacher and student networks within distillation frameworks, analyzing their effects on model performance. As shown in Table \ref{Strong-Student}, distillation is conducted using students with ResNet34 and ResNet50 backbones. Results indicate that the students’ AP$_{50}$ increases by 3.3\% and 3.5\%, respectively, even surpassing the teacher model. Moreover, the distilled students achieve 0.9\% and 1.5\% higher AP\textsubscript{75} than the teachers. The results demonstrate that our distillation method provides instructive guidance rather than forcing simple imitation. 
			In addition, to intuitively demonstrate the impact of knowledge distillation on model lightweighting, we provide Giga Floating Point Operations Per Second (GFLOPs), Parameters (Params), and Frames Per Second (FPS) performance in Table \ref{Strong-Student}. After distillation, the student surpasses the teacher in both lightweighting and accuracy.
			Furthermore, a stronger teacher with a ResNeXt101 backbone is employed. Table \ref{Strong-Teacher} reveals that stronger teachers fail to improve student performance and may even degrade it. We attribute this to the excessive performance gap that complicates knowledge transfer, as the teacher’s abstract representations hinder student comprehension.
			
		\end{enumerate*}
		
		\begin{table*}[tbp]
			\centering
			\caption{Experimental Results under the Stronger Student Configuration on DOTA Dataset}\label{Strong-Student}
			\begin{tblr}{
					hline{1,Z} = {1.5pt},
					colspec = {c|cc|cccccc|ccc},
				}
				Network & Method & Backbone & mAP & AP$_{50}$ & AP$_{75}$ & AP$_S$ & AP$_M$ & AP$_L$ & GFLOPs & Params/M & FPS\\
				\hline
				\SetCell[r=5]{m} RetinaNet & Teacher & ResNet101 & 36.5 & 60.4 & 37.4 & 11.8 & 37.6 & 48.6 & 293 & 55.6 & 16.4 \\
				\cline{2-12}
				& Student & ResNet34 & 33.8 & 57.1 & 34.8 & 11.4 & 35.3 & 43.5 & 202 & 30.2 & 26.7\\
				& Ours & ResNet34 & 36.6 & 60.4 (3.3$\uparrow$)& 38.3 & 12.8 & 37.9 & 48.2 & 202 & 30.2 & 26.7\\
				\cline{2-12}
				& Student & ResNet50 & 34.7 & 58.2 & 36.2 & 12.7 & 36.5 & 44.7 & 215 & 36.6 & 22.5\\
				& Ours & ResNet50 & 37.3 & 61.7 (3.5$\uparrow$)& 38.9 & 13.3 & 38.5 & 49.1 & 215 & 36.6 & 22.3\\
			\end{tblr}
		\end{table*}
		
		\begin{table}[tbp]
			\centering
			\caption{Experimental Results under the Stronger Teacher Configuration on DIOR Dataset}\label{Strong-Teacher}
			\begin{tblr}{
					hline{1,Z} = {1.5pt},
					colspec = {c|ccccc},
					colsep = 3pt,
				}
				Network & Method & Backbone & mAP & AP$_{50}$ & AP$_{75}$\\
				\hline
				\SetCell[r=3]{m} Faster R-CNN & Teacher & ResNeXt101 & 48.2 & 72.5 & 53.0\\
				& Student & ResNet18 & 41.9 & 67.1 & 45.0\\
				& Ours & ResNet18 & 45.7 & 70.8 (3.7$\uparrow$)& 49.7\\
				\hline
				\SetCell[r=3]{m} RetinaNet & Teacher & ResNeXt101 & 46.2 & 69.6 & 49.6\\
				& Student & ResNet18 & 38.8 & 62.9 & 40.7\\
				& Ours & ResNet18 & 42.6 & 66.1 (3.2$\uparrow$)& 45.6\\
			\end{tblr}
		\end{table}
		
		\begin{enumerate*}[font=\itshape, itemjoin=\\\hspace*{\parindent}]
		\setcounter{enumi}{5}
			\item \textit{Hyperparameter Tuning Experiments:} To verify algorithm robustness, hyperparameter adjustments are conducted. Specifically, we investigate the balancing effects between low-frequency and high-frequency losses: $L_{dist} = \gamma L_{low} + (2-\gamma)L_{high}$. Additionally, the balance between explicit and implicit losses is explored: $L_{dist} = \gamma L^{ex} + (2-\gamma)L^{im}$. Experimental results in Fig. \ref{LinePlot} depict the relationship between the balancing factor $\gamma$ (x-axis) and the AP$_{50}$ ratio (current/original ratio, y-axis). The original configuration ($\gamma=1$) achieves optimal performance. Performance variations under frequency balance adjustments demonstrate spectral decomposition’s essential role in optimally balancing low-frequency and high-frequency components. The minimal impact of stream balance adjustments confirms the algorithm’s robustness.
		\end{enumerate*}
		
		\begin{enumerate*}[font=\itshape, itemjoin=\\\hspace*{\parindent}]
			\setcounter{enumi}{6}		
			\item \textit{Impact of Wavelet Basis Selection:} To evaluate how different wavelet bases affect the distillation algorithm, we conducted comparative experiments using Haar, Daubechies-4, and Symlets-4 with Faster R-CNN. As shown in Table \ref{Ablation-Basis}, experimental results indicate minimal differences in DS\textsuperscript{2}D\textsuperscript{2}’s performance across these basis functions. For example, mAP values reached 45.6\%, 45.5\%, and 45.5\% for Haar, Daubechies-4, and Symlets-4, respectively. We attribute this consistency to the Dual-Stream design. It combines explicit and implicit distillation to extract valuable knowledge from spectral features, thereby reducing the dependence on specific wavelet bases.
			
		\end{enumerate*}
		
		\begin{table}[tbp]
			\centering
			\caption{Ablation Study of Different Wavelet Basis \\ Function on DIOR Dataset}\label{Ablation-Basis}
			\begin{tblr}{
					hline{1,Z} = {1.5pt},
					colspec = {c|cccccc},
				}
				Basis & mAP & AP$_{50}$ & AP$_{75}$ & AP$_S$ & AP$_M$ & AP$_L$\\
				\hline
				Haar & \textbf{45.6} & \textbf{70.9} & 49.7 & 11.9 & \textbf{34.2} & 63.4\\
				Daubechies & 45.5 & 70.5 & 49.6 & \textbf{12.1} & 34.1 & \textbf{63.8} \\
				Symlet & 45.5 & 70.4 & \textbf{49.9} & 11.1 & 33.7 & 63.6 \\
			\end{tblr}
		\end{table}
		
		\begin{enumerate*}[font=\itshape, itemjoin=\\\hspace*{\parindent}]
		\setcounter{enumi}{7}		
			\item \textit{Visualization of Overall Performance:} We visualize the student’s detection results before and after distillation to intuitively demonstrate algorithmic effectiveness. Fig. \ref{DIOR-Vis} presents Faster R-CNN’s detection results on DIOR dataset. Post-distillation performance is significantly improved, with the student model aligning closely with the ground truth in diverse scenarios. Our algorithm optimizes small objects (the first column) and dense objects (the third column), achieving marked enhancements. The original model struggles with aircraft detection under shadow interference, while our frequency-domain learning approach effectively circumvents such disturbances.
			
		\end{enumerate*}
		
\section{Conclusion}
	This paper proposes DS\textsuperscript{2}D\textsuperscript{2}, an efficient and versatile distillation framework for remote sensing object detection that integrates spectral features for explicit and implicit learning. DS\textsuperscript{2}D\textsuperscript{2} demonstrates the effectiveness of spectral knowledge in remote sensing data, which was previously overlooked by existing methods. Furthermore, we reveal and analyze the essential value of subtle feature map discrepancies in the distillation process. Therefore, DS\textsuperscript{2}D\textsuperscript{2} provides novel optimization insights for adapting distillation algorithms to remote sensing data. Through spectral analysis, DS\textsuperscript{2}D\textsuperscript{2} further demonstrates unique advantages in resisting interference and mitigating semantic confusion. Extensive experimental data substantiate the effectiveness and superiority of DS\textsuperscript{2}D\textsuperscript{2}. However, while the additional computational overhead introduced by the proposed method is inevitable, there may still be room for optimization. Future work will focus on extending distillation algorithms to multi-source remote sensing data interpretation applications.
\bibliographystyle{IEEEtran}
\bibliography{citepaper}

\vfill
\end{document}